\documentclass{article}
\usepackage[hidelinks]{hyperref}
\usepackage{array}
\usepackage{pgfplots}
\pgfplotsset{compat=1.18} 
\usepackage{spconf,amsmath,graphicx,hyperref}
\usepackage{afterpage}
\usepackage{booktabs}
\usepackage{multirow} 
\usepackage{hyperref}

\usepackage{amsmath,amssymb}
\usepackage{mathtools}
\usepackage{graphicx}   
\usepackage{xcolor}     
\usepgfplotslibrary{units} 
\usepackage{placeins}
\usepackage{subcaption} 

\usepgfplotslibrary{groupplots} 
\pgfplotsset{compat=1.18}       

\definecolor{PineGreen}{RGB}{0, 128, 0} 
\definecolor{brightyellow}{RGB}{204, 153, 0}
\definecolor{myBlue}{RGB}{68, 119, 170}
\definecolor{myGreen}{RGB}{143, 188, 143}
\definecolor{myOrange}{RGB}{244, 165, 130}
\definecolor{softorange}{RGB}{68, 119, 170}
\definecolor{aborange}{RGB}{255, 165, 113}
\definecolor{abgreen}{RGB}{165, 208, 141}
\definecolor{abpink}{RGB}{229, 124, 132}
\definecolor{introgray}{RGB}{231, 230, 230}
\definecolor{introblue}{RGB}{177, 226, 239}
\definecolor{intropink}{RGB}{247, 209, 239}

\newcommand{\stderr}[1]{\small $\pm$ #1}

\pgfplotsset{
    hyperparam_style/.style={
        ymin=50, ymax=90,
        ytick={50,55,60,65,70,75,80,85,90},
        ylabel={AUROC},
        axis lines=left,
        axis line style={draw=black!20},
        tick style={draw=none},
        grid=major,
        major grid style={draw=black!10, dashed},
        legend pos=north east,
        legend style={draw=none, fill=none, font=\small},
        y unit=\%,
        width=\linewidth, 
        height=4.5cm,      
    }
}

\title{Hallucination Detection {via} Internal States and Structured Reasoning Consistency in Large Language Models}
\name{Yusheng Song$^{1}$,
    Lirong Qiu$^{1}$,
    Xi Zhang$^{1}$,
    Zhihao Tang$^{1,\dagger}$\thanks{$^{\dagger}$Zhihao Tang is the corresponding author.}}
\address{$^{1}$ Beijing University of Posts and Telecommunications, Beijing, China \\
    \{songys, qiulirong, zhangx, innerone\}@bupt.edu.cn
    }
\begin{document}
\ninept
\maketitle

\afterpage{
    \begin{figure*}[t] 
    
        \centering
        \includegraphics[width=\textwidth]{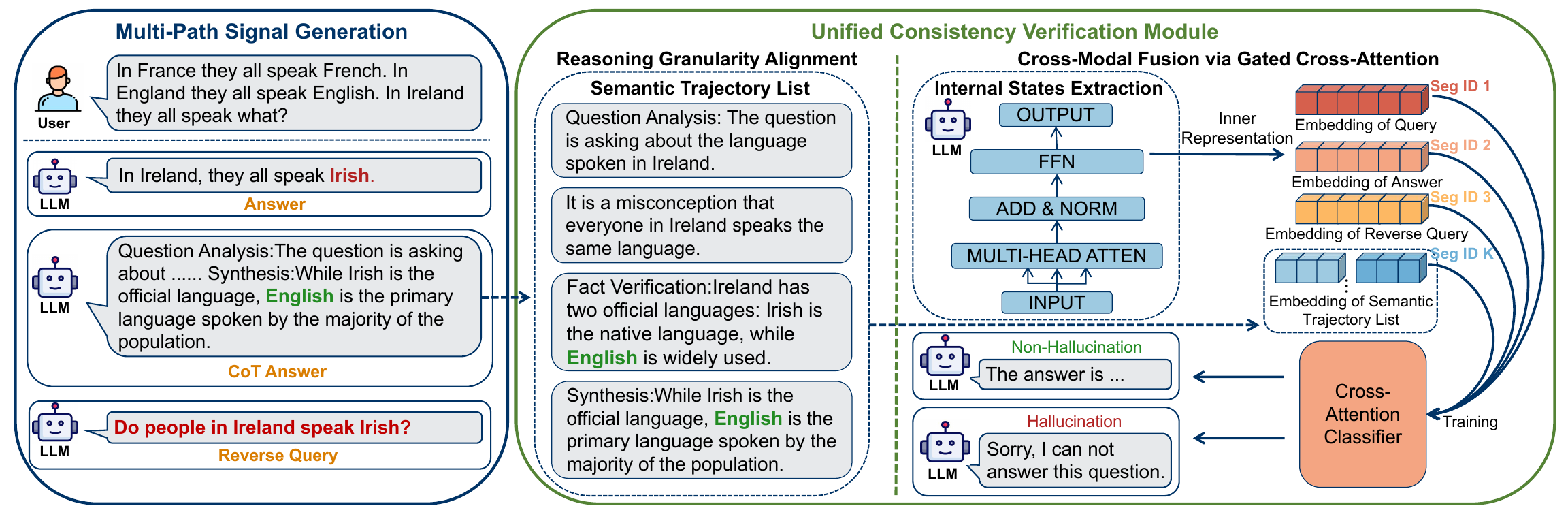} 
        \caption{Overview of the proposed hallucination detection framework.}
        \label{fig:main-picture}
        \vspace{-0.6cm}
    \end{figure*}
}

\begin{abstract}
\vspace{-0.1cm}

The detection of sophisticated hallucinations in Large Language Models (LLMs) is hampered by a ``Detection Dilemma'': methods probing internal states (Internal State Probing) excel at identifying factual inconsistencies but fail on logical fallacies, while those verifying externalized reasoning (Chain-of-Thought Verification) show the opposite behavior. This schism creates a task-dependent blind spot: Chain-of-Thought Verification fails on fact-intensive tasks like open-domain QA where reasoning is ungrounded, while Internal State Probing is ineffective on logic-intensive tasks like mathematical reasoning where models are confidently wrong. We resolve this with a unified framework that bridges this critical gap. However, unification is hindered by two fundamental challenges: the \textit{Signal Scarcity Barrier}, as coarse symbolic reasoning chains lack signals directly comparable to fine-grained internal states, and the \textit{Representational Alignment Barrier}, a deep-seated mismatch between their underlying semantic spaces. To overcome these, we introduce a multi-path reasoning mechanism to obtain more comparable, fine-grained signals, and a segment-aware temporalized cross-attention module to adaptively fuse these now-aligned representations, pinpointing subtle dissonances. Extensive experiments on three diverse benchmarks and two leading LLMs demonstrate that our framework consistently and significantly outperforms strong baselines. Our code is available: \href{https://github.com/peach918/HalluDet}{https://github.com/peach918/HalluDet}.

\end{abstract}

\vspace{-0.2cm}

\section{Introduction}
\vspace{-0.1cm}
\label{sec:intro}


Large Language Models (LLMs) are revolutionizing information interaction, yet a propensity to ``hallucinate''—generating plausible yet false content—critically undermines the technology's transformative potential~\cite{zhao2023survey}. 
Hallucination presents a fundamental flaw that challenges LLM reliability~\cite{kalai2025language}, especially in high-stakes domains like healthcare, finance, and law, where errors can lead to catastrophic outcomes~\cite{weidinger2021ethical},~\cite{huang2023survey}.
The resulting application bottleneck erodes systemic trust and prevents widespread adoption. 
Consequently, hallucination detection has become a cornerstone challenge for ensuring safe and trustworthy LLM applications.

\begin{figure}[t!]
\centering

\begin{subfigure}[t!]{0.55\linewidth}
    \centering
    \includegraphics[width=\linewidth]{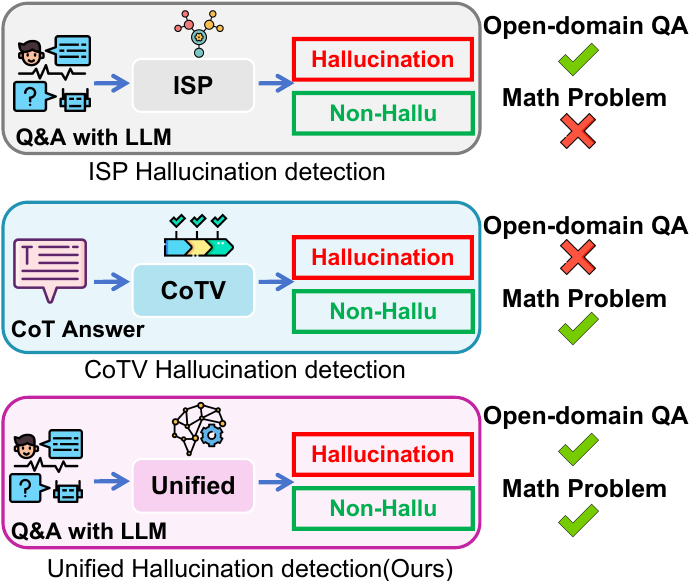}
    \caption{A visual comparison of the methods' effectiveness and performance.}
\end{subfigure}
\hspace{0.001cm}
\begin{subfigure}[t!]{0.43\linewidth}
    \flushleft
    \begin{tikzpicture}[scale=0.9]
      \begin{axis}[
        ybar,
        bar width=8pt,
        width=4.5cm, height=5cm,
        ylabel={AUROC (\%)},
        ylabel style={font=\fontsize{7pt}{6pt}\selectfont},
        yticklabel style={font=\footnotesize},
        symbolic x coords={Open-domain QA, Math Problem},
        xtick=data,
        xticklabel style={font=\fontsize{6.5pt}{11pt}\selectfont},
        ymin=60, ymax=90,
        legend style={at={(0.98,0.98)}, anchor=north east,font=\tiny, inner sep=1pt,cells={anchor=west}, area legend},
        legend image code/.code={%
          \fill[#1] (0cm,-0.1cm) rectangle (0.3cm,0.1cm);
        },
        ymajorgrids=true,
        grid style=dashed,
        enlarge x limits=0.40
      ]

        \addplot+[bar shift=-12pt, fill=introgray, draw=none] coordinates {
          (Open-domain QA,79.31)
          (Math Problem,70.36)
        };
        \addlegendentry{ISP}

        \addplot+[bar shift=0pt, fill=introblue, draw=none] coordinates {
          (Open-domain QA,63.91)
          (Math Problem,76.55)
        };
        \addlegendentry{CoTV}

        \addplot+[bar shift=12pt, fill=intropink, draw=none] coordinates {
          (Open-domain QA,84.03)
          (Math Problem,79.15)
        };
        \addlegendentry{Unified(ours)}

      \end{axis}
    \end{tikzpicture}
    \caption{Quantitative analysis of AUROC performance.}
\end{subfigure}

\caption{Effectiveness of detection methods for hallucination types.}
\label{fig:comparison}
\vspace{-0.5cm}
\end{figure}

Current efforts in hallucination detection are largely divided into two isolated paradigms.
First, the `neuroscientist's path' of Internal State Probing (ISP)~\cite{zhang2024transferable},~\cite{du2024haloscope} examines sub-symbolic signals within the model---such as neural activation patterns~\cite{su2024unsupervised}, token generation probabilities~\cite{quevedo2024detecting}, or semantic entropy~\cite{han2024semantic}---to find internal inconsistencies.
Second, the `psychologist's path' of Chain-of-Thought Verification (CoTV)~\cite{weng-etal-2023-large},~\cite{xue2023rcot} analyzes the logical coherence of the model's externalized, symbolic reasoning traces, often using self-verification protocols to detect contradictions~\cite{weng-etal-2023-large}, ~\cite{li2023making}.
However, these two paradigms have evolved not in concert, but largely in isolation. 
This ``binary schism'' in research is no accident; it reflects the long-standing methodological divide in artificial intelligence between sub-symbolic (connectionist) and symbolic (classicist) approaches~\cite{sun2001artificial}. 
This fracture creates a blind spot for detecting the most dangerous and subtle hallucinations.

This schism creates the \textit{Detection Dilemma}: a critical blind spot where each paradigm fails in a complementary manner. As illustrated in Fig.~\ref{fig:comparison}, the failures are task-dependent. ISP methods, while effective at gauging a model's statistical certainty, are blind to logical fallacies. They are thus ineffective in domains like mathematical reasoning, where a model can be highly confident in a logically flawed answer~\cite{beigi2024internalinspector}. Conversely, CoTV methods excel at verifying the internal coherence of a reasoning chain but cannot ground it in factual reality. They consequently fail in open-domain QA, where models build logical arguments on a factually incorrect premise, yielding self-consistent fabrications~\cite{hong2024closer}.
The essence of the Detection Dilemma is this decoupling of statistical confidence from factual grounding. Consequently, the most insidious hallucinations—those that are both statistically confident and logically coherent, yet factually baseless—evade detection by either method alone.

To resolve the Detection Dilemma, ISP and CoTV must be unified, yet this path is blocked by two fundamental technical challenges. 
The first is the \textbf{Signal Scarcity Barrier}. CoTV typically depends on a single reasoning path, which often appears logically self-consistent and thus fails to anchor sub-symbolic anomalies to concrete logical flaws. Consequently, anomalies detected by ISP cannot be validated against explicit reasoning, while CoTV evidence remains sparse. This lack of cross-paradigm indicators creates a semantic gap, yielding a scarcity of reliable hallucination signals.
The second is the \textbf{Representational Alignment Barrier}. Even when signals are encoded into a shared embedding format, their underlying semantic spaces remain heterogeneous. Embeddings of internal states capture latent statistical patterns, whereas embeddings of reasoning traces capture compositional logic. A direct vector comparison is therefore unreliable, confounded by a severe mismatch in both semantics and granularity (e.g., a fine-grained neural signal versus a coarse-grained reasoning step). Overcoming this alignment challenge is a critical prerequisite for successful unification.

A novel framework is introduced to resolve the Detection Dilemma by enforcing consistency between a model's internal states and its externalized reasoning. This is operationalized through two technical innovations.
First, to overcome the \textbf{Signal Scarcity Barrier}, a multi-path reasoning mechanism is employed to deliberately generate a diverse signal portfolio from both direct answers and auxiliary Chain-of-Thought (CoT). The CoT is then decomposed into a structured \textit{Semantic Trajectory List}. This critical step transforms the coarse symbolic trace into a fine-grained sequence, creating an explicit bridge that makes symbolic logic directly comparable to sub-symbolic neural states.
Second, to overcome the \textbf{Representational Alignment Barrier}, a segment-aware temporalized cross-Attention module is proposed. This component unifies the heterogeneous embeddings from questions, answers, and the now-structured CoT trajectories into a coherent representational space. By adaptively aligning these modalities, our module effectively detects the subtle semantic dissonances that are the hallmarks of sophisticated hallucinations. Extensive experiments on three public benchmarks validate our framework's effectiveness, consistently outperforming strong baselines. 

Our main contributions are summarized as follows:
\begin{itemize}
    \item We formally identify the Detection Dilemma in current research and propose the first unified framework to resolve it by bridging sub-symbolic and symbolic model representations.
    \item We introduce two technical innovations: a multi-path reasoning mechanism to address signal scarcity and a temporalized cross-attention to resolve representational misalignment.
    \item We demonstrate state-of-the-art performance on three diverse benchmarks, establishing a new standard for reliable hallucination detection.
\end{itemize}
\vspace{-0.4cm}

\afterpage{
\begin{figure*}[t!] 
    \centering
    \begin{subfigure}[b]{0.28\textwidth}
        \centering
        \begin{tikzpicture}
            \begin{axis}[
                hyperparam_style,
                width=\textwidth, 
                height=3.5cm,
                xmin=0, xmax=33,
                xtick={1,5,9,13,17,21,25,29,32},
                tick label style={font=\scriptsize},
                label style={font=\scriptsize},
                legend style={font=\scriptsize, yshift=1ex},
            ]
            \addplot[color=softorange, mark=*, mark options={fill=softorange, draw=softorange, scale=1.2}, smooth, line width=1.2pt] 
            coordinates { (1, 58.0) (5, 55.0) (9, 69.0) (13, 71.2) (17, 67.0) (21, 55.0) (25, 82.5) (29, 67.8) (32, 60.5) };
            \addlegendentry{Layer}
            \end{axis}
        \end{tikzpicture}
        \caption{Effect of Layer}
        \label{fig:sub_layer}
    \end{subfigure}
    \hspace{-0.4cm}
    \begin{subfigure}[b]{0.23\textwidth}
        \centering
        \begin{tikzpicture}
            \begin{axis}[
                hyperparam_style,
                ymin=65, ymax=90,
                xmin=0.1, xmax=0.9,
                width=\textwidth, 
                height=3.5cm,
                xtick={0.2,0.4,0.6,0.8},
                tick label style={font=\scriptsize},
                label style={font=\scriptsize},
                legend style={font=\scriptsize, yshift=1ex},
            ]
            \addplot[color=myGreen, mark=*, mark options={fill=myGreen, draw=myGreen, scale=1.2}, smooth, line width=1.2pt] 
            coordinates { (0.2, 68.76) (0.4, 76.36) (0.6, 72.78) (0.8, 82.42) };
            \addlegendentry{Temperature}
            \end{axis}
        \end{tikzpicture}
        \caption{Effect of Temperature}
        \label{fig:sub_temp}
    \end{subfigure}
    \hspace{-0.2cm}
    \begin{subfigure}[b]{0.24\textwidth}
        \centering
        \begin{tikzpicture}
            \begin{axis}[
                hyperparam_style,
                ymin=65, ymax=90,
                xmin=0, xmax=350,
                width=\textwidth, 
                height=3.5cm,
                xtick={50,100,200,300},
                tick label style={font=\scriptsize},
                label style={font=\scriptsize},
                legend style={font=\scriptsize, yshift=1ex},
            ]
            \addplot[color=myOrange, mark=*, mark options={fill=myOrange, draw=myOrange, scale=1.2}, smooth, line width=1.2pt] 
            coordinates { (50, 68.17) (100, 72.18) (200, 80.78) (300, 82.42) };
            \addlegendentry{Sequence Length}
            \end{axis}
        \end{tikzpicture}
        \caption{Effect of Sequence Length}
        \label{fig:sub_seqlen}
    \end{subfigure}
    \hspace{-0.2cm}
    \begin{subfigure}[b]{0.24\textwidth}
        \centering
        \begin{tikzpicture}
            \begin{axis}[
                hyperparam_style,
                ymin=65, ymax=90,
                xmin=0, xmax=34,
                width=\textwidth, 
                height=3.5cm,
                xtick={2,4,8,16,32},
                tick label style={font=\scriptsize},
                label style={font=\scriptsize},
                legend style={font=\scriptsize, yshift=1ex},
            ]
            \addplot[color=abpink, mark=*, mark options={fill=abpink, draw=abpink, scale=1.2}, smooth, line width=1.2pt] 
            coordinates { (2, 70.2) (4, 74.8) (8, 82.5) (16, 81.3) (32, 78.1) };
            \addlegendentry{Attention Heads}
            \end{axis}
        \end{tikzpicture}
        \caption{Effect of Attention Heads}
        \label{fig:sub_heads}
    \end{subfigure}
    \vspace{-0.1cm}
    \caption{Hyperparameter analysis for key parameters and their impact on AUROC.}
    \label{fig:hyperparam_analysis}
    \vspace{-0.6cm}
\end{figure*}
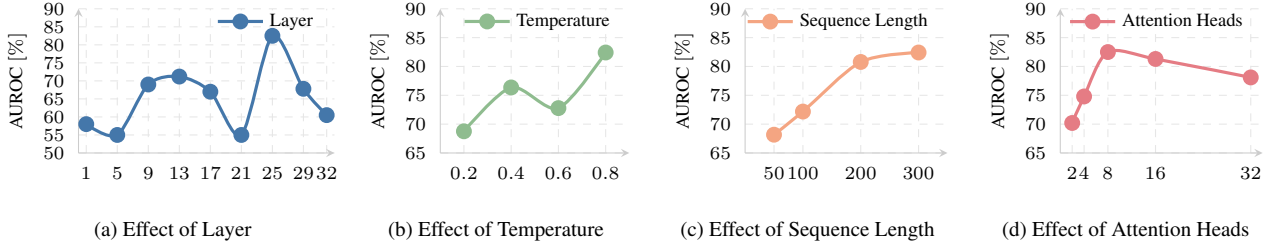
}

\section{Methodology}
\vspace{-0.1cm}
To resolve the `Detection Dilemma', we introduce a framework that integrates an LLM's internal sub-symbolic states with its externalized symbolic reasoning (Fig.~\ref{fig:main-picture}). Our design overcomes signal scarcity and representational misalignment via two innovations: a multi-path process to generate diverse signals, and a unified verification module to fuse them for discrepancy analysis. These components are detailed below.

\vspace{-0.3cm}

\subsection{Multi-Path Signal Generation for Comprehensive Diagnostics}
\vspace{-0.1cm}
\label{sec:signal_generation}

To overcome the signal scarcity barrier, our framework generates signals from three complementary reasoning paths for any input query $Q$, constructing a rich diagnostic landscape. This strategy performs cognitive triangulation, forcing the model to approach a problem from multiple angles to amplify latent inconsistencies indicative of hallucinations. The three paths are defined as follows:

\begin{enumerate}
    \item \textbf{Direct Answer Path:} The LLM is prompted to generate a direct answer, $A_{\text{dir}}$, without explicit intermediate reasoning. This path captures the model's spontaneous, unconditioned output, providing a baseline assessment of its immediate factual recall and statistical confidence.
    \item \textbf{Reasoning-Augmented Path:} The query Q is re-prompted with a Chain-of-Thought~\cite{wei2022chain} (CoT) instruction to elicit $A_{\text{cot}}$, a detailed response externalizing the model's step-by-step symbolic reasoning. This path renders the model's logical trajectory transparent and amenable to verification.
    \item \textbf{Reverse-Inference Path:} The direct answer $A_{\text{dir}}$ is supplied back to the LLM with the objective of inferring a plausible original query, $Q_{\text{rev}}$, that would logically lead to it. This path functions as a crucial semantic consistency check, probing whether the generated answer is sufficiently grounded to entail a question that aligns with the original query's intent.
\end{enumerate}

This tri-path generation strategy yields multi-perspective paired data ($Q$-$A_{\text{dir}}$, $Q$-$A_{\text{cot}}$, $A_{\text{dir}}$-$Q_{\text{rev}}$), which forms the foundation for our subsequent cross-modal consistency analysis.

For supervised training, high-quality hallucination labels are generated via a LLM-as-a-Judge protocol, which employs two state-of-the-art LLMs (GPT-4.1 and Gemini-2.5 Pro) to independently verify the target model's query–answer pairs. Each pair is assigned a binary label (0 for non-hallucination, 1 for hallucination). Pairs receiving concordant labels from both judges are incorporated directly into our dataset. Domain experts manually resolve disagreements to efficiently produce a large-scale, highly accurate labeled dataset.

\vspace{-0.25cm}

\subsection{Unified Consistency Verification Module}
\vspace{-0.1cm}
\label{sec:consistency_verification}

The core technical engine of our framework is a unified verification module engineered to resolve the representational alignment barrier. This module addresses the complex challenge of comparing and integrating heterogeneous signals—namely, the sub-symbolic hidden states from the LLM's neural pathways and the symbolic, structured text from the reasoning-augmented path. Its architecture follows a two-stage process: first, aligning the granularity of the symbolic reasoning trace with internal state representations, and second, merging them to detect semantic and logical dissonances.

\vspace{-0.2cm}

\subsubsection{Reasoning Granularity Alignment via Temporal Modeling}
\vspace{-0.1cm}

To align the variable-length symbolic CoT response $A_{\text{cot}}$ with fixed-dimensional neural representations, 
a semantic trajectory decomposition is first performed. Specifically, $A_{\text{cot}}$ is segmented into a sequence of minimal, coherent semantic units ${u_1, u_2, \dots, u_m}$, termed the Semantic Trajectory List (STL). This segmentation follows linguistic cues like logical connectors (e.g., ``therefore'', ``because''), causal transitions, and fact-introduction points to maintain the integrity of each reasoning step. The resulting STL is then subjected to temporal embedding, enabling structured and fine-grained modeling of the reasoning process in a neural representation space.


The resulting STL offers a structured representation of the reasoning process, defined as a sequence of embeddings:

\vspace{-0.2cm}
\begin{equation}
    T = [e_1, e_2, \dots, e_m],
\end{equation}

where $e_i = \text{Enc}(u_i)$ is the embedding of the $i$-th reasoning unit, obtained from the same encoder as the answer-generating LLM. This clause-level representation captures both the temporal progression of the logic and the fine-grained semantic shifts between steps.

To distill the sequential information into a compact representation, temporal modeling is employed.
Specifically, a learnable classification token \texttt{[CLS]} is prepended to the sequence of trajectory embeddings and processed with a Transformer encoder~\cite{vaswani2017attention}. This architecture is chosen for its ability to capture long-range dependencies, essential for reasoning chains where a conclusion depends on a distant premise. The final aggregated representation of the CoT path, $h_{\text{CoT}}$, is extracted from the output state of the token:

\vspace{-0.2cm}
\begin{equation}
h_{\text{CoT}} = \text{Enc}([\texttt{[CLS]}; e_1, \dots, e_T])_{\texttt{[CLS]}}.
\end{equation}

This procedure performs semantic compression, creating a holistic vector that encapsulates the entire reasoning trajectory while aligning its granularity with other internal state representations.

\vspace{-0.2cm}

\subsubsection{Cross-Modal Fusion via Gated Cross-Attention}
\vspace{-0.1cm}

With all signals transformed into a shared representational format, the final stage involves their integration and analysis to detect inconsistencies. This fusion process follows a hierarchical verification: first ensuring consistency within the sub-symbolic domain, then performing a cross-modal check against the symbolic reasoning trace.

\noindent \textbf{Internal State Extraction and Contextualization.}
For the non-CoT paths ($Q$, $A_{\text{dir}}$, $Q_{\text{rev}}$), their vector representations are obtained in two ways. For $A_{\text{dir}}$ and $Q_{\text{rev}}$, the corresponding hidden states are directly extracted from the LLM during generation. For $Q$, as an input rather than a generated output, it is fed into the same LLM, with its embedding layer used to produce $E_Q$. To preserve the origin and role of each representation, a unique Segment ID is assigned to each embedding (e.g., one for queries, another for answers).
Subsequently, $E_Q$, $E_{A_{\text{dir}}}$, and $E_{Q_{\text{rev}}}$ are concatenated into a sequence $X_{\text{main}}$, which is passed through a Multi-Head self-Attention~\cite{vaswani2017attention} (MHA) block to perform an intra-modal consistency check, yielding $H_{\text{main}}$ that captures relationships and discrepancies among these signals.

\vspace{-0.2cm}
\begin{equation}
    H_{\text{main}} = \text{MHA}(\text{LayerNorm}(X_{\text{main}} + E_{\text{seg}})).
\end{equation}

\noindent \textbf{Adaptive Reasoning Gate.}
To dynamically regulate the influence of the symbolic reasoning path, a gating mechanism is introduced. A scalar gate $g \in \mathbb{R}$ is computed from the contextualized internal states $H_{\text{main}}$ and applied to modulate the CoT representation $h_{\text{CoT}}$. This enables the model to down-weight the reasoning trace if internal signals deem it unreliable or irrelevant for a given instance.

\vspace{-0.2cm}
\begin{equation}
    g = \sigma(\text{FFN}(H_{\text{main}})), \quad \hat{h}_{\text{CoT}} = g \cdot h_{\text{CoT}}.
\end{equation}

\noindent \textbf{Cross-Attention for Discrepancy Detection.}
The core of our verification is a final inter-modal consistency check, implemented through a cross-attention~\cite{chen2021crossvit} module. The contextualized internal states $H_{\text{main}}$ serve as a set of queries to probe the gated symbolic reasoning representation $\hat{h}_{\text{CoT}}$, which provides the key-value context. The output, $Z$, is a fused representation where dissonances between the model's sub-symbolic ``knowledge'' and symbolic ``explanation'' are highlighted by the attention mechanism.

\vspace{-0.2cm}
\begin{equation}
    Z = \text{CrossAttn}(H_{\text{main}}, \hat{h}_{\text{CoT}}).
\end{equation}
This fused representation $Z$ is then passed through a final MLP classifier to produce the logits $l \in \mathbb{R}^2$ for hallucination prediction.

\noindent \textbf{Optimization.}
Due to the natural class imbalance between hallucinated and factual statements, the model is optimized using Focal Loss~\cite{lin2017focal} ($L_{\text{FL}}$), which prioritizes hard-to-classify examples:

\vspace{-0.2cm}
\begin{equation}
    L_{\text{FL}} = -\alpha_t(1 - p_t)^\gamma \log(p_t),
\end{equation}
where $p_t$ is the model's estimated probability for the ground-truth class, $\gamma$ is a focusing parameter, and $\alpha_t$ is a weighting factor to balance class importance. This cross-attention-based fusion architecture enables unified, end-to-end modeling that captures subtle yet critical hallucination signals by identifying patterns of disagreement across different modalities of the model's own cognitive processes.

\vspace{-0.2cm}

\begin{table}[t!]
\centering
\caption{Main hallucination detection results. Best baseline results are \underline{underlined}. Gains of our method are highlighted in \textcolor{PineGreen}{green}.}
\label{tab:main_table}
\resizebox{\columnwidth}{!}{%
\begin{tabular}{llccc}
\toprule
\textbf{LLM} & \textbf{Method} & \textbf{TruthfulQA} & \textbf{TriviaQA} & \textbf{GSM8K} \\
\midrule
\multirow{7}{*}{Qwen2.5-7B} & SAPLMA~\cite{azaria2023internal} & 59.66~\stderr{1.69} & 62.36~\stderr{1.38} & 59.72~\stderr{1.91} \\
 & selfcheckgpt~\cite{manakul2023selfcheckgpt} & 55.08~\stderr{1.15} & 74.65~\stderr{0.92} & 67.98~\stderr{1.28} \\
 & semantic entropy~\cite{han2024semantic} & 64.72~\stderr{1.26} & 75.68~\stderr{0.88} & 58.36~\stderr{1.46} \\
 & V-STaR~\cite{hosseini2024v} & 63.91~\stderr{0.93} & 71.09~\stderr{1.15} & \underline{76.55~\stderr{1.21}} \\
 & HaloScope~\cite{du2024haloscope} & \underline{79.31~\stderr{2.33}} & \underline{81.52~\stderr{2.08}} & 70.36~\stderr{2.46} \\
 & \textbf{ours} & \textbf{84.03}~\stderr{1.69} & \textbf{85.68}~\stderr{1.71} & \textbf{79.15}~\stderr{1.68} \\
 \cmidrule(lr){2-5}
 & \textit{Gain vs. Best} & \textcolor{PineGreen}{+4.72} & \textcolor{PineGreen}{+4.16} & \textcolor{PineGreen}{+2.60} \\
\midrule
\multirow{7}{*}{Llama2-7B-chat} & SAPLMA~\cite{azaria2023internal} & 57.41~\stderr{1.71} & 60.32~\stderr{1.83} & 58.64~\stderr{2.02} \\
 & selfcheckgpt~\cite{manakul2023selfcheckgpt} & 52.95~\stderr{1.20} & 73.22~\stderr{1.01} & 62.36~\stderr{1.35} 
\\
 & semantic entropy~\cite{han2024semantic} & 62.17~\stderr{1.32} & 73.65~\stderr{0.77} & 57.46~\stderr{1.53} \\
 & V-STaR~\cite{hosseini2024v} & 61.28~\stderr{1.12} & 68.83~\stderr{1.33} & \underline{74.38~\stderr{1.25}} \\
 & HaloScope~\cite{du2024haloscope} & \underline{78.64~\stderr{2.25}} & \underline{77.40~\stderr{1.98}} & 65.79~\stderr{2.31} \\
 & \textbf{ours} & \textbf{82.42}~\stderr{1.63} & \textbf{79.46}~\stderr{1.87} & \textbf{76.83}~\stderr{2.06} \\
 \cmidrule(lr){2-5}
 & \textit{Gain vs. Best} & \textcolor{PineGreen}{+3.78} & \textcolor{PineGreen}{+2.06} & \textcolor{PineGreen}{+2.45} \\
\bottomrule
\end{tabular}%
}
\vspace{-0.65cm}
\end{table}

\section{Experiments}

\subsection{Experimental Setup}
\vspace{-0.1cm}

To ensure methodological rigor, we evaluate on two distinct LLMs, LLaMA2-7B-Chat~\cite{touvron2023llama} and Qwen2.5-7B~\cite{qwen2024qwen2}, to demonstrate generalizability. Our testbed embodies the ``Detection Dilemma'' using three benchmarks: fact-intensive \textbf{TruthfulQA}~\cite{lin2022truthfulqa}, logic-intensive \textbf{GSM8K}~\cite{cobbe2021training}, and \textbf{TriviaQA}~\cite{joshi2017triviaqa}. We compare against state-of-the-art ISP (HaloScope~\cite{du2024haloscope}, SAPLMA~\cite{azaria2023internal}) and CoTV (V-STaR~\cite{hosseini2024v}) baselines. Performance is measured by AUROC, with all decoding at a fixed temperature of 0.8 and a maximum length of 300 tokens.

\vspace{-0.2cm}

\subsection{Quantitative Performance Comparison}
\vspace{-0.1cm}

As presented in Table~\ref{tab:main_table}, our unified framework consistently and significantly outperforms all baselines, providing robust empirical evidence for its superiority in resolving the ``Detection Dilemma''. This dilemma is empirically manifested in the specialized performance of prior methods: the CoTV-based V-STaR~\cite{hosseini2024v} excels on logic-intensive tasks (GSM8K: 76.55\%) but fails on fact-intensive ones (TruthfulQA: 63.91\%), while the ISP-based HaloScope~\cite{du2024haloscope} exhibits the opposite trade-off (GSM8K: 70.36\% vs. TruthfulQA: 79.31\%). Our framework breaks this trade-off, achieving state-of-the-art performance on both TruthfulQA (84.03\%) and GSM8K (79.15\%) simultaneously. This balanced, high-level performance across fundamentally different tasks demonstrates that by successfully unifying sub-symbolic and symbolic signals, our approach overcomes prior weaknesses to achieve a more generalized detection capability.

\vspace{-0.2cm}

\subsection{In-depth Analysis}
\vspace{-0.1cm}

To analyze the framework's performance, a series of analyses was conducted to explain not just \textit{what} it achieves, but \textit{why} it is effective.

\begin{figure}[t!]
    \centering
    \includegraphics[width=0.85\linewidth]{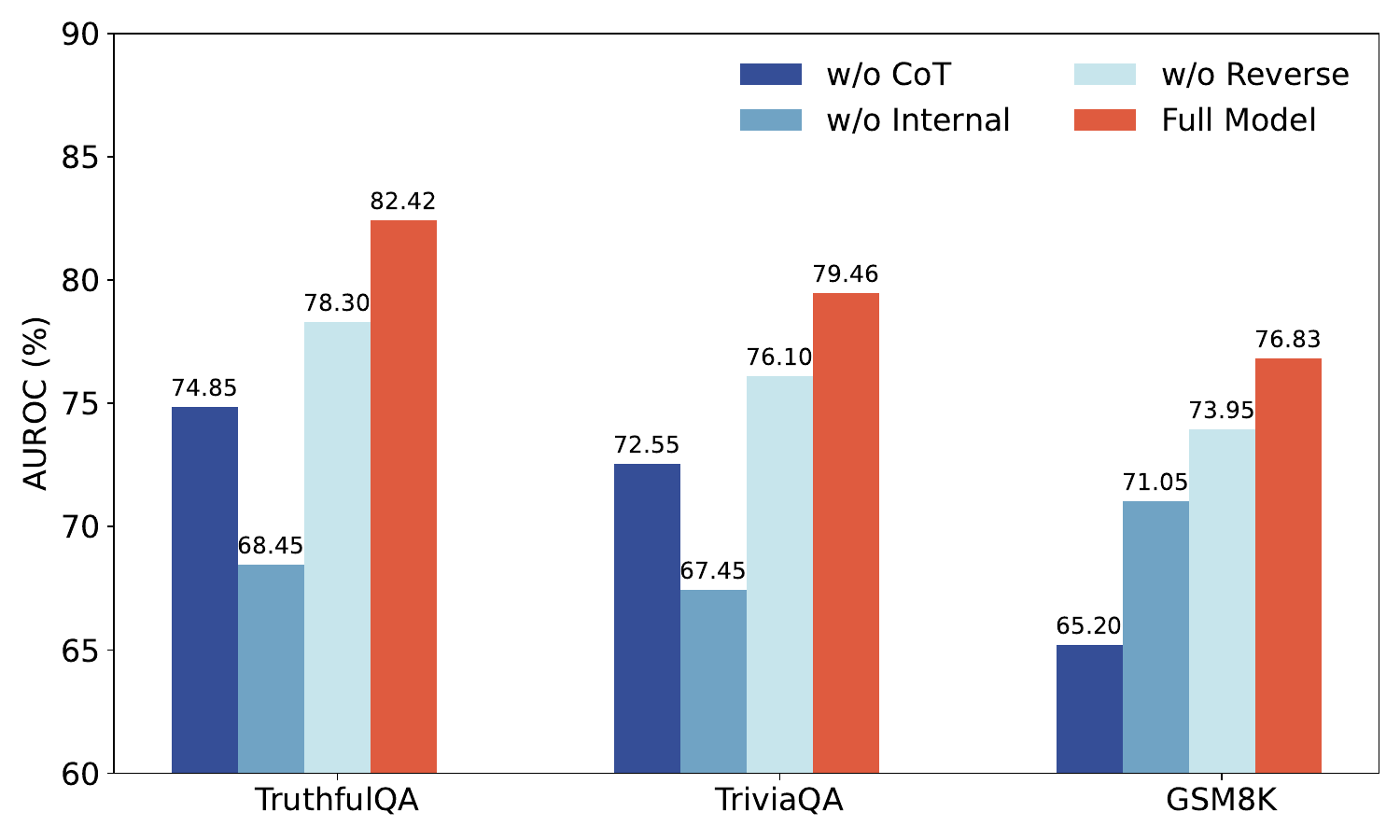}
    \caption{Ablation Study:Internal States, CoT \& Reverse Inference.}
    \label{fig:Ablation}
    \vspace{-0.5cm}
\end{figure}

\noindent\textbf{Component-wise Efficacy.} An ablation study (Fig.~\ref{fig:Ablation}) reveals a strong synergy between our primary innovations. Removing Internal States (w/o Internal), the CoT verification path (w/o CoT), or the Reverse Inference Path (w/o Reverse) degrades performance, and the full model’s improvement is non-linear. On TruthfulQA, our full model achieves a 4.12-point AUROC improvement over the best component. This suggests our cross-attention fusion deeply integrates the signals—CoT for transparent reasoning, Internal States for statistical patterns, and Reverse Inference for semantic consistency. This synergy is key to resolving the Detection Dilemma.



\begin{figure}[b]
\centering
\includegraphics[width=0.85\linewidth]{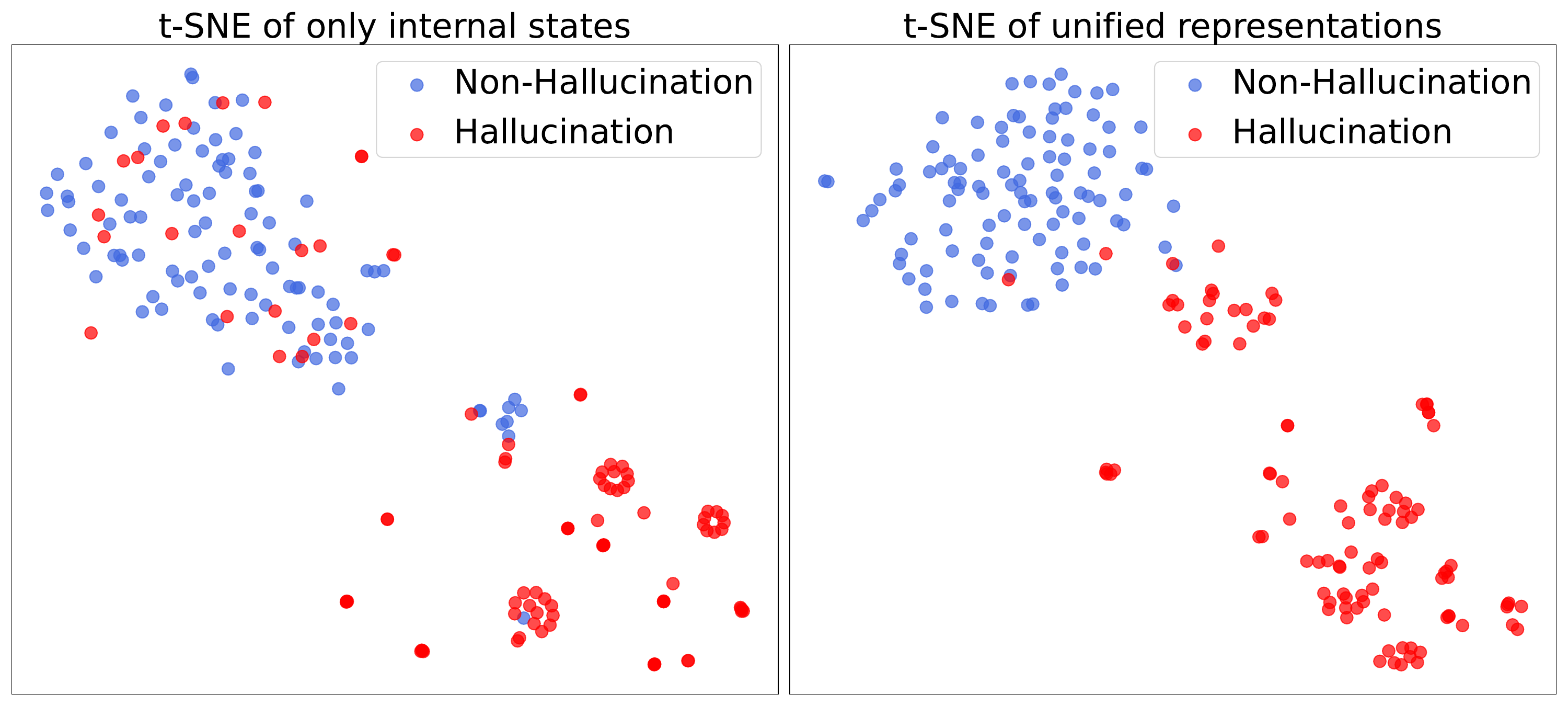} 
\caption{Comparison of t-SNE projections. The visualization distinguishes hallucination (\textcolor{red}{red}) from non-hallucination (\textcolor{blue}{blue}) samples.} 
\label{fig:tsne}
\vspace{-0.3cm}
\end{figure}


\begin{figure}[h]
\centering
\includegraphics[width=0.85\linewidth]{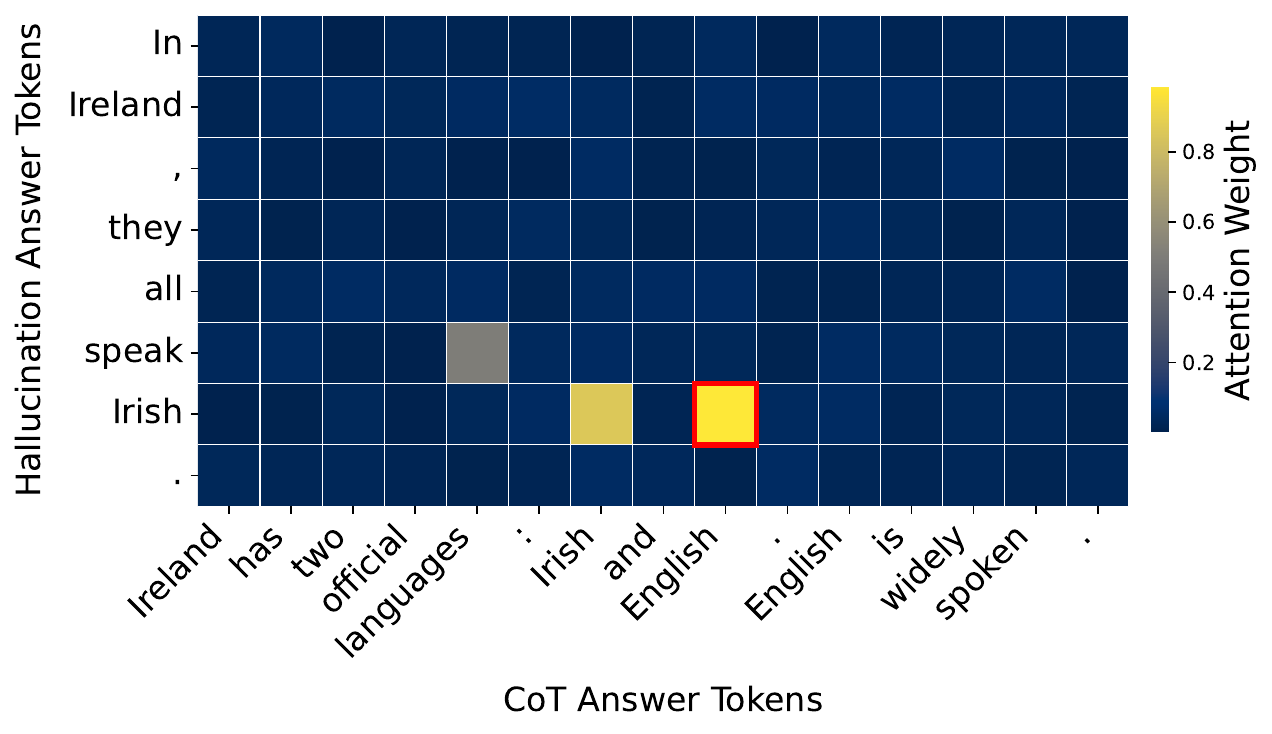} 
\vspace{-0.3cm}
\caption{Visualization of cross-attention weights between a hallucinated answer and the corresponding CoT trace.
} 
\vspace{-0.65cm}
\label{fig:attention}
\end{figure}

\noindent\textbf{Model Characteristics and Interpretability.} Hyperparameter analysis (Fig.~\ref{fig:hyperparam_analysis}) provides further insight into the model's operational logic. Optimal performance is achieved using late-stage representations (24th layer), confirming that hallucination detection relies on abstract semantic features. The framework's accuracy is highest for text generated at a higher temperature ($T=0.8$), indicating it is most effective in the creative (and thus higher-risk) scenarios where it is most needed. To provide intuitive visual evidence of the framework's mechanics, qualitative analyses were performed. A t-SNE~\cite{maaten2008visualizing} projection of the feature space (Fig.~\ref{fig:tsne}) shows that our unified representations achieve significantly better separation between hallucinated (\textcolor{red}{red}) and non-hallucinated (\textcolor{blue}{blue}) samples compared to using only internal states, demonstrating superior discriminative power. Furthermore, visualizing the cross-attention weights (Fig.~\ref{fig:attention}) provides interpretability. In the given example, the model correctly places high attention on tokens that create a semantic dissonance between a false statement (``In Ireland they all speak Irish'') and corrective facts in the CoT (``two official languages,'' ``English spoken widely''). This confirms that the framework's decisions are grounded in identifiable semantic contradictions rather than opaque correlations, enhancing trust in its predictions.

\vspace{-0.3cm}

\section{CONCLUSION}
\vspace{-0.1cm}
\label{sec:typestyle}

In this work, we address the ``Detection Dilemma'' in LLM hallucination, a vulnerability from the schism between Internal State Probing and Chain-of-Thought Verification, by introducing the first unified framework to bridge sub-symbolic and symbolic signals. Our approach overcomes the \textit{Signal Scarcity Barrier} with a multi-path reasoning mechanism and the \textit{Representational Alignment Barrier} via a segment-aware temporalized cross-Attention module. Significant performance gains across diverse benchmarks validate our thesis that this synergistic approach is essential for robust detection, representing a critical step towards building trustworthy LLMs for high-stakes applications.

\bibliographystyle{IEEEbib}
\bibliography{strings,refs}

\end{document}